# Anti-drift in electronic nose via dimensionality reduction: a discriminative subspace projection approach

**Zhengkun Yi[1], Cheng Li[2]**
[1]School of Computing, National University of Singapore.
[2]Shenzhen Institutes of Advanced Technology, Chinese Academy of Sciences.

E-mail: cheng.li6@siat.ac.cn

**Abstract.** Sensor drift is a well-known issue in the field of sensors and measurement and has plagued the sensor community for many years. In this paper, we propose a sensor drift correction method to deal with the sensor drift problem. Specifically, we propose a discriminative subspace projection approach for sensor drift reduction in electronic noses. The proposed method inherits the merits of the subspace projection method called domain regularized component analysis. Moreover, the proposed method takes the source data label information into consideration, which minimizes the within-class variance of the projected source samples and at the same time maximizes the between-class variance. The label information is exploited to avoid overlapping of samples with different labels in the subspace. Experiments on two sensor drift datasets have shown the effectiveness of the proposed approach.

## 1. Introduction

Electronic noses (E-noses) have been widely used in a wealth of domains, for instance, indoor and outdoor air quality monitoring [1,2], medical diagnosis [3,4], detection of polluting gases from vehicles [5,6], and fruit quality control [7]. An electronic nose device is commonly comprised of a sensor array, a conditioning circuit, and a signal processing electronic system [8]. In the past decades, tremendous efforts have been made to develop various gas sensors based on different sensing principles [9]. For instance, a single generic tin oxide gas sensor was reported for the first time to discriminate among complex odors [10]. Meanwhile, sensor conditioning circuits have also been improved for better gas sensing. As an example, a compact and low-cost electronic circuit was developed by Flammini et al. [11]. This device is capable of supporting a wide range of resistive values, which is key to the realization of an electronic nose.

E-nose as a gas sensing device suffers from the sensor drift issue, which is a well-known problem in the field of sensors and measurement [12,13] and has plagued the sensor community for many years. Sensor drift can be generally divided into two categories [14]. One category is called the first-order drift caused by aging and poisoning, and the other category is called the second-order drift caused by uncontrollable alterations of the experimental operating system such as temperature and humidity variations. In practical applications, it is difficult to clearly distinguish these two types of sensor drift.

Drift compensation or drift correction can be implemented in different aspects such as hardware updating [15] and signal improvement. In the signal processing level, approaches of drift correction are broadly divided into three categories, i.e., univariate methods, multivariate methods, and machine learning approaches [16]. Typical univariate methods include frequency analysis, baseline manipulation, and differential measurements [13]. The univariate methods correct the response of each sensor independently, which are simple but extremely sensitive to sample rate variations [17,18]. Unlike the univariate methods, multivariate methods correct the responses of the entire sensor array [19,20]. Both the univariate methods and multivariate methods explicitly do the sensor drift compensation. However, it may be impossible to compensate the sensor drift in many real-world applications. Therefore, machine learning methods are employed to implicitly do the sensor drift correction via learning from data distributions. In this paper, we focus on developing machine learning approaches for sensor drift compensation.

The main contribution of this paper is to propose a discriminative subspace projection method for drift reduction in electronic noses. The proposed method inherits the merits of the subspace projection method called domain regularization component analysis (DCRA) in [21]. Moreover, the proposed method takes the source data label information into consideration, which minimizes the within-class variance of projected source samples and at the same time maximizes the between-class variance. The label information is exploited to avoid overlapping of samples with different labels in the subspace. Figure 1 illustrates the basic idea of the proposed discriminative domain regularization component analysis (D-DCRA).

The rest of the paper is organized as follows. Section 2 briefly provides the related works on drift correction in electronic noses and relevant machine learning approaches. Subsequently, Section 3 presents the proposed discriminative subspace projection. In section 4, the results of our method and the competing methods on two gas sensor datasets are compared. Finally, Section 5 concludes the work and discusses the future work.

## 2. Related work

In this section, the machine learning methods for anti-drift in electronic noses are firstly reviewed in Section 2.1. Subsequently, Section 2.2 provides a brief review of subspace learning methods. In Section 2.3 we present another quite related machine learning topic, i.e., domain adaptation, which is widely used in many applications in addition to sensor drift correction.

### 2.1 Anti-drift in E-nose using machine learning methods

Topic of anti-drift in E-nose using machine learning methods is receiving increasing attention in the past several decades. Vergara et al. [14] contributed an extensive gas sensor drift dataset including six gases with concentration ranging from 10 to 1000 ppmv. The dataset was collected over a period of three years using an array of 16 metal-oxide gas sensor. In addition, classifier ensembles as a drift compensation tool were employed to solve the gas recognition problem. Zhang et al. [22] proposed two drift compensation algorithms based on extreme learning machines [23]. The proposed domain adaptation extreme learning machine achieved the gas classification by leveraging a limited number of labelled samples from target domain. The computational efficiency of domain adaptation extreme learning machine was comparable to that of classic extreme learning machines. Ziyatdinov et al. [24] proposed a drift reduction method called common principal component analysis (CPCA). This approach addressed the sensor drift issue by searching for components that are common for all gases. Zhang et al. [21] presented an unsupervised subspace learning approach for drift reduction. The proposed method aimed to find a subspace by minimizing the mean distribution discrepancy. The distribution difference between the source domain and target domain was very small in the subspace. The subspace search could be easily solved by eigenvalue decomposition.

## 2.2 Dimensionality reduction methods via subspace projection

The objective of subspace learning is to find a low-dimensional subspace through optimizing certain objective functions. Principal component analysis (PCA) is an unsupervised subspace learning method [25]. PCA is an orthogonal linear transformation that converts a number of possibly correlated variables into a number of relatively small uncorrelated variables. The data are projected along the directions of maximal variance. Each principal component accounts for as much of the variability of the original data as possible. Locality preserving projection (LPP) is another unsupervised method on the manifold [26]. LPP optimally preserves the neighborhood structure of the data set and is able to discover the nonlinear structure of the data manifold. Compared with PCA and LPP, Linear discriminant analysis (LDA) [27] is a supervised dimensionality reduction method, which maximizes the between-class variance and at the same time minimize the within-class variance. A graph embedding method was proposed as a general framework for dimensionality reduction called Marginal Fisher Analysis (MFA) [28]. MFA is also a supervised subspace approach, which overcome the limitations of LDA by designing two graphs to characterize the within-class compactness and between-class separability.

## 2.3 Domain adaptation

Classic machine learning approaches assume that the distribution of training set is consistent with that of test set. However, the distribution discrepancy between the training set and test set is a common issue in various real-world applications such as text classification [29], sentiment analysis [30], cross-system recommendation [31], and indoor WiFi localization [32]. Transfer learning is a machine learning technique proposed to alleviate the distribution discrepancy issue [33]. Domain adaptation [34,35] is a branch of transfer learning, which aims to improve the algorithm performance in the target domain by utilizing the information from both source and target domain. Pan et al. [36] proposed a domain adaptation method called transfer component analysis (TCA) by minimizing the Maximum Mean Discrepancy (MMD) [37]. Transfer components span a subspace in which data distributions of the source domain and target domain are close to each other. TCA is further extended in a semi-supervised learning setting [36]. Jiang et al. [38] presented an algorithm called integration of global and local metrics for domain adaptation learning (IGLDA). Unlike TCA, IGLDA take the source data label information into consideration as well as minimizing MMD. Classifier and feature-invariant subspace are commonly learned independently in domain adaptation problem. Long et al. proposed a unified framework to achieve both distribution adaptation and label propagation named Adaptation Regularization based Transfer Learning (ARTL) [39]. Table 1 provides the list of abbreviations used in this paper.

## 3. The proposed discriminative domain regularized component analysis (D-DRCA)

In this section, the notation employed throughout this paper is given in Section 3.1. Subsequently, Section 3.2 provides a brief introduction of domain regularized component analysis. Section 3.3 presents the proposed discriminative domain regularized component analysis, which is an improved extension of DRCA.

### 3.1 Notations

Table 2 provides the list of notations employed in this paper. The sample set of source domain is denoted as $\mathbf{X}_s = [\mathbf{x}_s^1, \cdots, \mathbf{x}_s^{N_s}] \in \mathbb{R}^{D \times N_s}$ and the sample set of target domain is denoted as $\mathbf{X}_t = [\mathbf{x}_t^1, \cdots, \mathbf{x}_t^{N_t}] \in \mathbb{R}^{D \times N_t}$. $N_s$ and $N_t$ are the number of samples in source domain and target domain, respectively. $D$ is the dimension of original space and $d$ is the dimension of the subspace. The projection matrix is represented as $\mathbf{P} \in \mathbb{R}^{D \times d}$. The projected samples in the source domain is given by

$$\mathbf{Y}_s = [\mathbf{y}_s^1, \cdots, \mathbf{y}_s^{N_s}] = \mathbf{P}^T \mathbf{X}_s = \mathbf{P}^T [\mathbf{x}_s^1, \cdots, \mathbf{x}_s^{N_s}] \in \mathbb{R}^{d \times N_s}, \tag{1}$$

and the projected samples in the target domain is given by

$$\mathbf{Y}_t = [\mathbf{y}_t^1, \cdots, \mathbf{y}_t^{N_t}] = \mathbf{P}^T \mathbf{X}_t = \mathbf{P}^T [\mathbf{x}_t^1, \cdots, \mathbf{x}_t^{N_t}] \in \mathbb{R}^{d \times N_t}. \tag{2}$$

Table 2 list the notations defined in this paper.

### 3.2 Domain regularized component analysis (DRCA)

Domain regularized component analysis [21] is an unsupervised method without using any data label information. The main idea of DRCA is to learn a projection matrix such that the projected sample set of source domain has similar probability distribution as that of target domain. The projection matrix is computed by optimizing two terms simultaneously including 1) minimize the mean distribution discrepancy (MDD) between the project sample set $\mathbf{Y}_s$ and the project sample set $\mathbf{Y}_t$ and 2) maximize the variance (i.e., energy) of both source data and target data.

**Mean distribution discrepancy (MDD):** The mean distribution discrepancy is defined as the distance between the mean of projected samples of source domain $\overline{\mathbf{y}_s} = \frac{1}{N_s} \sum_{i=1}^{N_s} \mathbf{y}_s^i$ and the mean of projected samples of target domain $\overline{\mathbf{y}_t} = \frac{1}{N_t} \sum_{i=1}^{N_t} \mathbf{y}_t^i$. The minimization of mean distribution discrepancy is given by

$$\begin{aligned}
\min_{\mathbf{P}} \|\overline{\mathbf{y}_s} - \overline{\mathbf{y}_t}\|_2^2 &= \min_{\mathbf{P}} \left\| \frac{1}{N_s} \sum_{i=1}^{N_s} \mathbf{y}_s^i - \frac{1}{N_t} \sum_{i=1}^{N_t} \mathbf{y}_t^i \right\|_2^2 \\
&= \min_{\mathbf{P}} \left\| \frac{1}{N_s} \sum_{i=1}^{N_s} \mathbf{P}^T \mathbf{x}_s^i - \frac{1}{N_t} \sum_{i=1}^{N_t} \mathbf{P}^T \mathbf{x}_t^i \right\|_2^2 \\
&= \min_{\mathbf{P}} \|\mathbf{P}^T \overline{\mathbf{x}_s} - \mathbf{P}^T \overline{\mathbf{x}_t}\|_2^2 \\
&= \min_{\mathbf{P}} Tr[\mathbf{P}^T (\overline{\mathbf{x}_s} - \overline{\mathbf{x}_t})(\overline{\mathbf{x}_s} - \overline{\mathbf{x}_t})^T \mathbf{P}].
\end{aligned} \tag{3}$$

**Variance of projected samples:** The maximization of variance of projected source samples is given by

$$\max_{\mathbf{P}} Tr(\mathbf{Y}_s \mathbf{Y}_s^T) = \max_{\mathbf{P}} Tr[\mathbf{P}^T (\mathbf{X}_s \mathbf{X}_s^T) \mathbf{P}]. \tag{4}$$

Similarly, the maximization of variance of projected target samples is given by

$$\max_{\mathbf{P}} Tr(\mathbf{Y}_t \mathbf{Y}_t^T) = \max_{\mathbf{P}} Tr[\mathbf{P}^T (\mathbf{X}_t \mathbf{X}_t^T) \mathbf{P}]. \tag{5}$$

The DRCA incorporates the mean distribution discrepancy term and variance term together, which can be formulated as

$$\max_{\mathbf{P}} \frac{Tr[\mathbf{P}^T(\mathbf{X}_s\mathbf{X}_s^T + \lambda\mathbf{X}_t\mathbf{X}_t^T)\mathbf{P}]}{Tr[\mathbf{P}^T(\overline{\mathbf{x}_s} - \overline{\mathbf{x}_t})(\overline{\mathbf{x}_s} - \overline{\mathbf{x}_t})^T\mathbf{P}]}, \tag{6}$$

where $\lambda$ denotes the trade-off parameter to avoid bias learning between source and target domain. The problem can be easily solved by eigen decomposition.

### 3.3 Discriminative domain regularized component analysis (D-DRCA)

The proposed discriminative domain regularized component analysis inherits the merits of DRCA. i.e., D-DRCA is also to optimize the MDD term and the variance term. The MDD term can be considered as a global metric [38]. However, DRCA ignores the data label information in the source domain. It is desirable to make the projected samples with different labels more discriminative. Therefore, D-DRCE takes the label information into consideration by optimizing two additional terms including 1) minimize within-class distance of projected samples in source domain and 2) maximize between-class distance of projected samples in source domain.

Let $c$ be the number of labels and $n_l$ is the number of samples of the class $l$ ($1 \leq l \leq c$) in the source domain. The sample set of the $l$-th class in the source domain are denoted as $\mathbf{X}_s^l = [\mathbf{x}_s^{l1}, \cdots, \mathbf{x}_s^{ln_l}]$. The projected sample set of the $l$-th class in the source domain are denoted as $\mathbf{Y}_s^l = [\mathbf{y}_s^{l1}, \cdots, \mathbf{y}_s^{ln_l}]$.

**Within-class distance of projected samples in source domain:** The minimization of the within-class distance of projected samples in source domain is defined by

$$\min_{\mathbf{P}} \sum_{l=1}^{c} \sum_{j=1}^{n_l} \left\| \mathbf{y}_s^{lj} - \overline{\mathbf{y}_s^l} \right\|_2^2$$
$$= \min_{\mathbf{P}} \sum_{l=1}^{c} \sum_{j=1}^{n_l} \left\| \mathbf{P}^T\mathbf{x}_s^{lj} - \mathbf{P}^T\overline{\mathbf{x}_s^l} \right\|_2^2$$
$$= \min_{\mathbf{P}} \sum_{l=1}^{c} \sum_{j=1}^{n_l} Tr\left[ \mathbf{P}^T\left(\mathbf{x}_s^{lj} - \overline{\mathbf{x}_s^l}\right)\left(\mathbf{x}_s^{lj} - \overline{\mathbf{x}_s^l}\right)^T \mathbf{P} \right] \tag{7}$$

where $\overline{\mathbf{x}_s^l}$ denotes the mean of the samples of the $l$-th class in the source domain and $\overline{\mathbf{y}_s^l}$ denotes the mean of the projected samples of the $l$-th class in the source domain.

**Between-class distance of projected samples in source domain:** The maximization of the between-class distance of projected samples in source domain is given by

$$\max_{\mathbf{P}} \sum_{l=1}^{c} n_l \left\| \overline{\mathbf{y}_s^l} - \overline{\mathbf{y}_s} \right\|_2^2$$
$$= \max_{\mathbf{P}} \sum_{l=1}^{c} n_l \left\| \mathbf{P}^T\overline{\mathbf{x}_s^l} - \mathbf{P}^T\overline{\mathbf{x}_s} \right\|_2^2$$
$$= \max_{\mathbf{P}} \sum_{l=1}^{c} n_l \times Tr\left[ \mathbf{P}^T\left(\overline{\mathbf{x}_s^l} - \overline{\mathbf{x}_s}\right)\left(\overline{\mathbf{x}_s^l} - \overline{\mathbf{x}_s}\right)^T \mathbf{P} \right]. \tag{8}$$

Given both the within-class distance and the between-class distance, D-DRCA extends DRCA with the formulation given by

$$\max_{\mathbf{P}} \frac{Tr\left[\mathbf{P}^T\left(\frac{1}{N_s}\mathbf{X}_s\mathbf{X}_s^T + \lambda\frac{1}{N_t}\mathbf{X}_t\mathbf{X}_t^T - \kappa\sum_{l=1}^{c}\sum_{j=1}^{n_l}\frac{1}{cn_l}\left(\mathbf{x}_s^{lj} - \overline{\mathbf{x}_s^l}\right)\left(\mathbf{x}_s^{lj} - \overline{\mathbf{x}_s^l}\right)^T + \mu\sum_{l=1}^{c}\frac{n_l}{c}\left(\overline{\mathbf{x}_s^l} - \overline{\mathbf{x}_s}\right)\left(\overline{\mathbf{x}_s^l} - \overline{\mathbf{x}_s}\right)^T\right)\mathbf{P}\right]}{Tr[\mathbf{P}^T(\overline{\mathbf{x}_s} - \overline{\mathbf{x}_t})(\overline{\mathbf{x}_s} - \overline{\mathbf{x}_t})^T\mathbf{P}]} \tag{9}$$

where $\lambda$, $\kappa$, and $\mu$ are the trade-off parameters. To make parameters easily tunable, each term in the numerator in equation (9) in normalized.

The solution of equation (9) is not unique [21]. The problem is reformulated in the following to guarantee a unique solution

$$\max_{\mathbf{P}} Tr\left[\mathbf{P}^T\left(\frac{1}{N_s}\mathbf{X}_s\mathbf{X}_s^T + \lambda\frac{1}{N_t}\mathbf{X}_t\mathbf{X}_t^T - \kappa D_{wc} + \mu D_{bc}\right)\mathbf{P}\right]$$

$$s.t.\ Tr[\mathbf{P}^T(\overline{\mathbf{x}_s} - \overline{\mathbf{x}_t})(\overline{\mathbf{x}_s} - \overline{\mathbf{x}_t})^T\mathbf{P}] = \rho \tag{10}$$

where $\rho$ is a positive constant, $D_{wc} = \sum_{l=1}^{c}\sum_{j=1}^{n_l}\frac{1}{cn_l}\left(\mathbf{x}_s^{lj} - \overline{\mathbf{x}_s^l}\right)\left(\mathbf{x}_s^{lj} - \overline{\mathbf{x}_s^l}\right)^T$ and $D_{bc} = \sum_{l=1}^{c}\frac{n_l}{c}\left(\overline{\mathbf{x}_s^l} - \overline{\mathbf{x}_s}\right)\left(\overline{\mathbf{x}_s^l} - \overline{\mathbf{x}_s}\right)^T$.

Lagrangian method is employed to solve the optimization problem is equation (10). The Lagrangian $L(\mathbf{P}, \rho)$ associated with equation (10) is given by

$$L(\mathbf{P}, \rho) = Tr\left[\mathbf{P}^T\left(\frac{1}{N_s}\mathbf{X}_s\mathbf{X}_s^T + \lambda\frac{1}{N_t}\mathbf{X}_t\mathbf{X}_t^T - \kappa D_{wc} + \mu D_{bc}\right)\mathbf{P}\right] - \rho Tr[\mathbf{P}^T(\overline{\mathbf{x}_s} - \overline{\mathbf{x}_t})(\overline{\mathbf{x}_s} - \overline{\mathbf{x}_t})^T\mathbf{P}]. \tag{11}$$

By $\partial L(\mathbf{P}, \rho)/\partial\mathbf{P} = 0$, we have

$$\left((\overline{\mathbf{x}_s} - \overline{\mathbf{x}_t})(\overline{\mathbf{x}_s} - \overline{\mathbf{x}_t})^T\right)^{-1}\left(\frac{1}{N_s}\mathbf{X}_s\mathbf{X}_s^T + \lambda\frac{1}{N_t}\mathbf{X}_t\mathbf{X}_t^T - \kappa D_{wc} + \mu D_{bc}\right)\mathbf{P} = \eta\mathbf{P}. \tag{12}$$

Therefore, $\mathbf{P}$ can be obtained via eigenvalue decomposition. The optimal projection matrix is given by

$$\mathbf{P}^* = [\mathbf{p}_1, \cdots \mathbf{p}_d] \tag{13}$$

where $\mathbf{p}_i\ (1 \le i \le d)$ are the eigenvectors corresponding to the first $d$ largest eigenvalues. The proposed D-DRCA algorithm is depicted in Figure 2.

## 4. Results and discussion

In this section, experiments are performed to demonstrate the effectiveness of the proposed approach on two sensor drift datasets including one form UCSD and the other from CQU. the proposed discriminative domain regularized component analysis is compared with DCRA and other competing methods.

### 4.1 Experiment on sensor drift dataset from UCSD

The sensor drift dataset from UCSD was collected by Vergara et al [14]. A total of 13910 samples were collected using an electronic nose consisting of 16 gas sensors. The collection period was 36 months from January 2018 to February 2011. There were totally six types of gases at different concentrations to be detected, which include Acetaldehyde, Acetone, Ammonia, Ethanol, Ethylene, and Toluene. All the samples were split into ten batches based on the sample acquisition time. The detailed information regarding the acquisition time and the sample count of each batch is summarized in Table 3. Feature extraction were performed for each sensor. The feature number of each sensor is 8, resulting in 128-dimensional feature vector for each sample.

Following the experimental protocol in [16] and [14], batch 1 is adopted as the samples in the source domain with label information. Other batches are adopted as the samples in the target domains whose labels need to be predicted. Figure 3 depicts the 2D projection of the samples in batch 1~10. The time-varying sensor drift can be obviously observed, i.e., the distribution difference between the source domain and target domain is time-dependent.

There are 13 competing methods in total. PCA and LDA are baseline subspace approaches. Component correction based principal component analysis (CC-PCA) is a multivariate method [20]. Multi-class support vector machine with RBF kernel (SVM-rbf), the geodesic flow kernel (SVM-gfk), and the combination kernel (SVM-comgfk) are methods presented in [16]. ML-rbf and ML-comfgk are semi-supervised methods with manifold regularization [16]. Orthogonal signal correction (OSC) is another multivariate method similar to CC-PCA, which aims to find the undesired component through searching for subspace that is orthogonal to the target variable [40]. Both generalized least squares weighting (GLSW) [41] and direct standardization (DS) [42] are calibration transfer methods. DRCA is a recent subspace method proposed by Zhang et al [21].

The recognition accuracy of the sensor drift dataset from UCSD is shown in Table 4. The proposed D-DRCA algorithm outperforms the competing methods in terms of average classification accuracy achieving the highest value of 73.80%. Moreover, the proposed D-DRCA performs best in four drift correction tasks, i.e., batch 1→ batch 3, batch 1→ batch 5, batch 1→ batch 6, and batch 1→ batch 10.

### 4.2 Experiment on sensor drift dataset from CQU

The sensor drift dataset from CQU was collected by Zhang et al [21]. A total of 1604 samples were collected using multiple E-nose devices of the same model. Therefore, the sensor drift might be caused by device differences in this dataset. The dataset is comprised of three batches, which includes batch master collected five years earlier than the batches slave 1 and slave 2. There were totally six types of gases to be detected, which include Ammonia, Benzene, Carbon monoxide, Formaldehyde, Nitrogen dioxide and Toluene. The detailed information regarding the sample count of each batch is summarized in Table 5. Feature extraction were performed for each sensor resulting in 6-dimensional feature vector for each sample.

Following the experimental protocol in [21], master is adopted as the samples in the source domain with label information. Other batches including slave 1 and slave 2 are adopted as the samples in the target domains whose labels need to be predicted. Figure 4 depicts the 2D projection of the samples in batch master, slave 1, and slave 2 respectively. The time-varying sensor drift can be obviously observed, i.e., the distribution difference between the source domain and target domain is time-dependent.

There are 6 competing methods in total. Specifically, the competing methods contain SVM, PCA, LDA, calibration transfer methods (GLSW and DS) and DRCA. The recognition accuracy of the sensor drift dataset from CQU is shown in Table 6. The proposed D-DRCA method outperforms the competing ones in terms of average classification accuracy achieving the highest value of 65.23%. Moreover, the proposed D-DRCA performs best in both individual tasks, i.e., master → slave 1, and master → slave 2.

### 4.3 Parameter sensitivity analysis

There are four parameters to be tuned in the proposed D-DRCA method. Specifically, the parameters contain the dimension of subspace $d$, regularization coefficient $\lambda$, within-class coefficient $\kappa$, and between-class coefficient $\mu$. The parameter $d$ is tuned from the set $\{2^k, k = 0, 1, 2, 3, 4, 5, 6, 7\}$ for the case of the UCSD dataset and from the set $\{1, 2, 3, 4, 5, 6, 7\}$ for the case of the CQU dataset. For both datasets, the regularization coefficient $\lambda$, within-class coefficient $\kappa$, and between-class coefficient $\mu$, are all tuned from the set $\{10^k, k = -2, -1, 0, 1, 2\}$. Figure 5 shows the classification accuracy of the proposed D-DRCA method on UCSD dataset by tuning the dimension of subspace $d$ and regularization coefficient $\lambda$. Figure 6 shows the classification accuracy of the proposed D-DRCA method on UCSD dataset by tuning within-class coefficient $\kappa$, and between-class coefficient $\mu$. Similarly, Figure 7 shows the classification accuracy of the proposed D-DRCA method on CQU dataset by tuning the dimension of subspace $d$ and regularization coefficient $\lambda$. Figure 8 shows the classification accuracy of the proposed D-DRCA method on CQU dataset by tuning within-class coefficient $\kappa$, and between-class coefficient $\mu$.

### 5. Conclusion

In this paper, we propose a discriminative domain regularized component analysis (D-DCRA) approach for sensor drift compensation problem. The proposed method is inspired by machine learning approaches including domain adaptation and linear discriminant analysis. D-DCRA has the advantages of the previous method called DCRA, such as, it can be easily solved by eigenvalue decomposition. In addition, the proposed approach takes the label information in the source domain into account as well. The exploitation of label information can avoid overlapping of the samples with different labels in the subspace. The effectiveness of the proposed approaches has been verified on two public sensor drift datasets. Future work could introduce the kernel method to reduce high-order distribution difference between the source and target domain. Another potential direction is to employ graph embedding method to generate the domain-invariant features.

**Acknowledgement**

The authors would like to thank Dr. Vergara's group from USCD and Dr. Zhang's group for providing the sensor drift datasets. The authors would also thank Dr. Zhang for his clear and helpful explanation on the DCRA method.

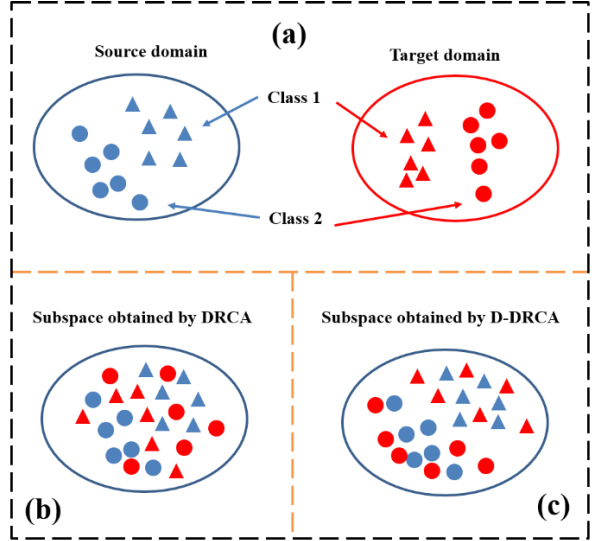

**Figure 1.** The illustration of discriminative domain regularization component analysis. (a) samples in source and target domain with different distributions. (b) The distribution difference between source and target is alleviated by DRCA, but samples with different labels overlapping in the subspace. (c) D-DRCA method reduce the distribution difference and avoid the overlapping problem simultaneously.

---

**Algorithm 1.** The proposed discriminative domain regularized component analysis (D-DRCA)

1: **Input:**
  - The sample set in source domain $\mathbf{X}_s$ ;
  - The sample set in target domain $\mathbf{X}_t$ ;
2: Compute the mean of samples in source domain $\overline{\mathbf{x}_s}$ and the mean of samples in target domain $\overline{\mathbf{x}_t}$;
3: Compute the within-class distance $D_{wc}$;
4: Compute the between-class distance $D_{bc}$;
5: Solve the eigenvalue decomposition problem in equation (13)
6: **Output**
  - The optimal projection matrix $\mathbf{P}^* = [\mathbf{p}_1, \cdots, \mathbf{p}_d]$ where $\mathbf{p}_i (1 \leq i \leq d)$ are the eigenvectors corresponding to the first $d$ largest eigenvalues of equation (13);
  - The projected sample set in source domain $\mathbf{Y}_s = (\mathbf{P}^*)^T \mathbf{X}_s$;
  - The projected sample set in target domain $\mathbf{Y}_t = (\mathbf{P}^*)^T \mathbf{X}_t$.

---

**Figure 2.** The proposed discriminative domain regularized component analysis (D-DRCA) algorithm

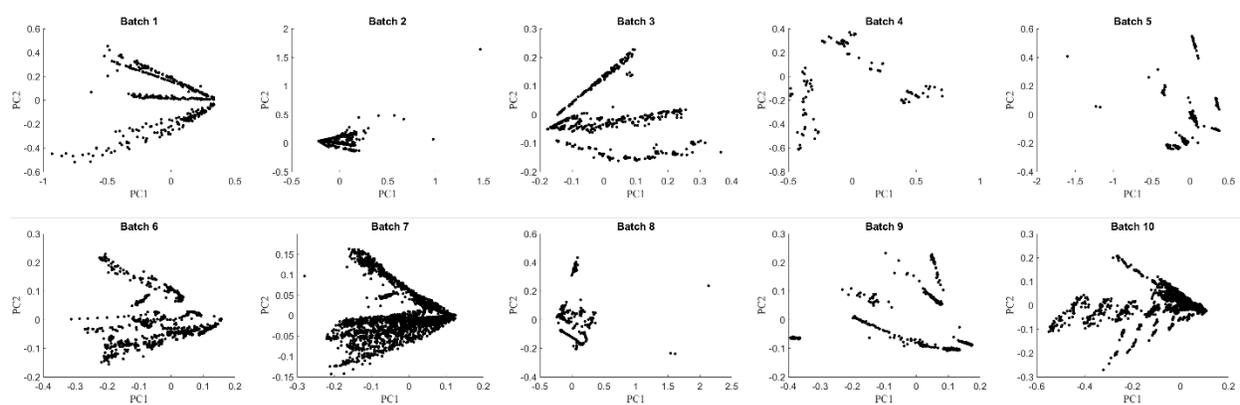

**Figure 3.** Samples in batch 1~10 are projected to 2D subspace using principal component analysis. The time-varying sensor drift can be easily observed.

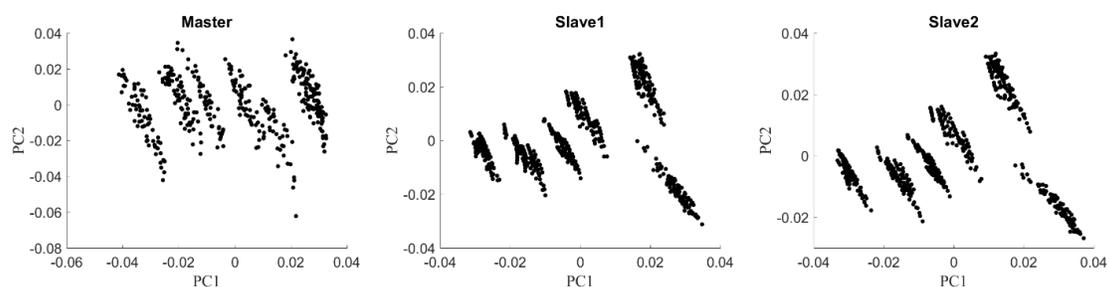

**Figure 4.** Samples in batch master, slave 1 and slave 2 are projected to 2D subspace using principal component analysis respectively. The time-varying sensor drift can be easily observed.

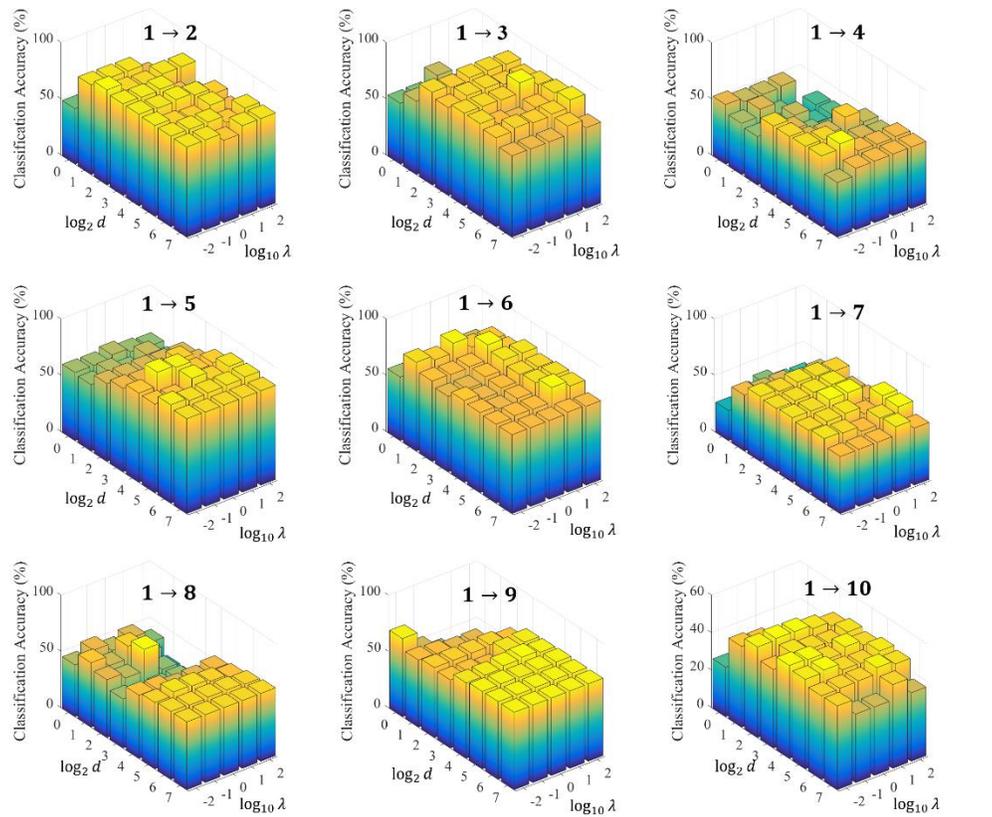

**Figure 5.** Classification accuracy of the proposed D-DRCA method on UCSD dataset by tuning the dimension of subspace $d$ and regularization coefficient $\lambda$.

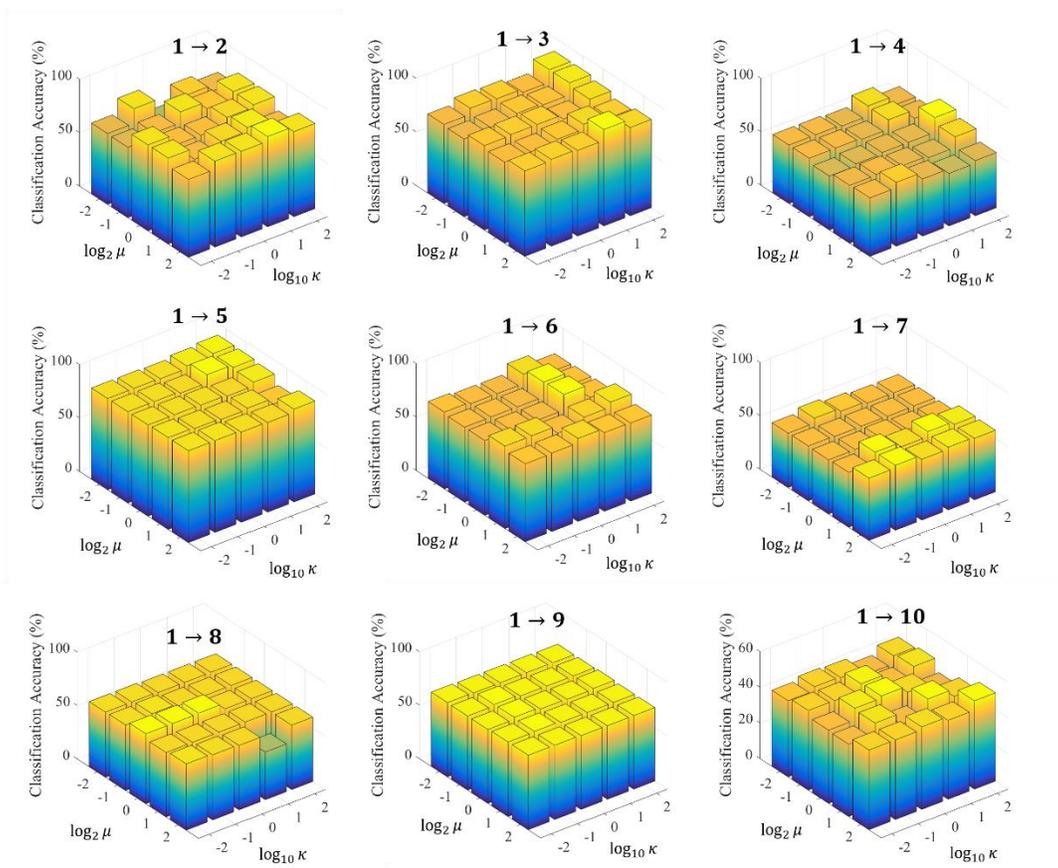

**Figure 6.** Classification accuracy of the proposed D-DRCA method on UCSD dataset by tuning within-class coefficient $\kappa$, and between-class coefficient $\mu$.

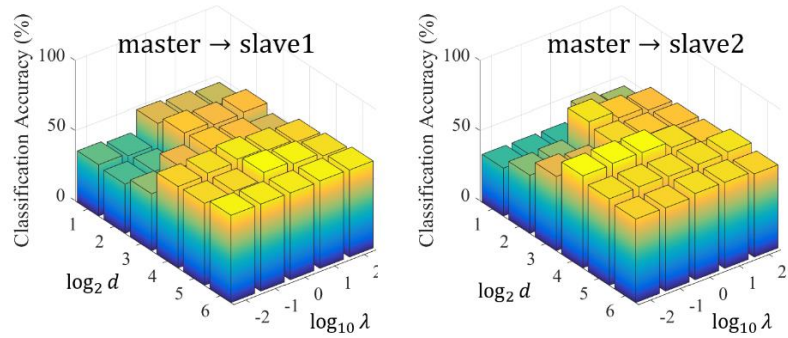

**Figure 7.** Classification accuracy of the proposed D-DRCA method on CQU dataset by tuning the dimension of subspace $d$ and regularization coefficient $\lambda$.

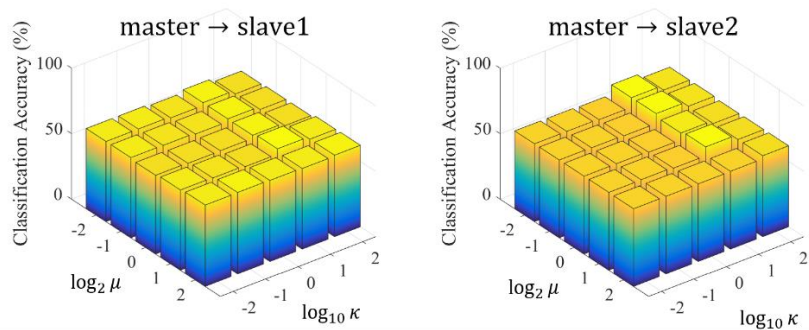

**Figure 8.** Classification accuracy of the proposed D-DRCA method on CQU dataset by tuning within-class coefficient $\kappa$, and between-class coefficient $\mu$.

**Table 1.** List of abbreviations.

| | |
|---|---|
| E-nose | electronic nose |
| LDA | linear discriminant analysis |
| LPP | locality preserving projection |
| PCA | principal component analysis |
| CPCA | common principal component analysis |
| CC-PCA | component correction based principal component analysis |
| OSC | orthogonal signal correction |
| GLSW | generalized least squares weighting |
| DS | direct standardization |
| MFA | marginal fisher analysis |
| MDD | mean distribution discrepancy |
| DCRA | domain regularized component analysis |
| D-DCRA | discriminative domain regularized component analysis |
| SVM | support vector machine |
| TCA | transfer component analysis |
| MMD | maximum mean discrepancy |
| IGLDA | integration of global and local metrics for domain adaptation learning |
| ARTL | adaptation regularization based transfer learning |

**Table 2.** List of notations.

| | |
|---|---|
| $\mathbf{X}_s$ | the sample set in source domain |
| $\mathbf{X}_t$ | the sample set in target domain |
| $\overline{\mathbf{x}_s}$ | the mean of samples in source domain |
| $\overline{\mathbf{x}_t}$ | the mean of samples in target domain |
| $\mathbf{x}_s^i$ | the $i$-th sample of the sample set in source domain |
| $\mathbf{x}_t^i$ | the $i$-th sample of the sample set in target domain |
| $\mathbf{P}$ | the projection matrix |
| $\mathbf{Y}_s$ | the projected sample set in source domain |
| $\mathbf{Y}_t$ | the projected sample set in target domain |
| $\overline{\mathbf{y}_s}$ | the mean of the projected samples in source domain |
| $\overline{\mathbf{y}_t}$ | the mean of the projected samples in target domain |
| $\mathbf{y}_s^i$ | the $i$-th sample of the projected sample set in source domain |
| $\mathbf{y}_t^i$ | the $i$-th sample of the projected sample set in target domain |
| $N_s$ | the number of samples in source domain |
| $N_t$ | the number of samples in target domain |
| $n_l$ | the number of samples of the class $l$ in source domain |
| $\mathbf{X}_s^l$ | the sample set of the $l$-th class in source domain |
| $\mathbf{Y}_s^l$ | The projected sample set of the $l$-th class in source domain |
| $\mathbf{x}_s^{li}$ | the $i$-th sample of the sample set of the $l$-th class in source domain |
| $\mathbf{y}_s^{li}$ | the $i$-th sample of the projected sample set of the $l$-th class in source domain |
| $\overline{\mathbf{x}_s^l}$ | the mean of the samples of the $l$-th class in source domain |
| $\overline{\mathbf{y}_s^l}$ | the mean of of the projected sample set of the $l$-th class in source domain |
| $c$ | the number of labels |

**Table 3**. Gas sensor drift dataset from UCSD with period of collection and sample count of different types of gases.

| Batch ID | Month | Ethanol | Ethylene | Ammonia | Acetaldehyde | Acetone | Toluene | Total |
|---|---|---|---|---|---|---|---|---|
| 1 | 1, 2 | 90 | 98 | 83 | 30 | 70 | 74 | 445 |
| 2 | 3, 4, 8~10 | 164 | 334 | 100 | 109 | 532 | 5 | 1244 |
| 3 | 11~13 | 365 | 490 | 216 | 240 | 275 | 0 | 1586 |
| 4 | 14, 15 | 64 | 43 | 12 | 30 | 12 | 0 | 161 |
| 5 | 16 | 28 | 40 | 20 | 46 | 63 | 0 | 197 |
| 6 | 17~20 | 514 | 574 | 110 | 29 | 606 | 467 | 2300 |
| 7 | 21 | 649 | 662 | 360 | 744 | 630 | 568 | 3613 |
| 8 | 22, 23 | 30 | 30 | 40 | 33 | 143 | 18 | 294 |
| 9 | 24, 30 | 61 | 55 | 100 | 75 | 78 | 101 | 470 |
| 10 | 36 | 600 | 600 | 600 | 600 | 600 | 600 | 3600 |

**Table 4.** Recognition accuracy (%) of the sensor drift dataset from UCSD. Bold values represent the best results. The proposed D-DRCA algorithm outperforms the competing methods on average.

| Batch ID | Batch 2 | 3 | 4 | 5 | 6 | 7 | 8 | 9 | 10 | Average |
|---|---|---|---|---|---|---|---|---|---|---|
| PCA$_{SVM}$ | 82.40 | 84.80 | 80.12 | 75.13 | 73.57 | 56.16 | 48.64 | 67.45 | 49.14 | 68.60 |
| LDA$_{SVM}$ | 47.27 | 57.76 | 50.93 | 62.44 | 41.48 | 37.42 | **68.37** | 52.34 | 31.17 | 49.91 |
| CC-PCA | 67.00 | 48.50 | 41.00 | 35.50 | 55.00 | 31.00 | 56.50 | 46.50 | 30.50 | 45.72 |
| SVM-rbf | 74.36 | 61.03 | 50.93 | 18.27 | 28.26 | 28.81 | 20.07 | 34.26 | 34.47 | 38.94 |
| SVM-gfk | 72.75 | 70.08 | 60.75 | 75.08 | 73.82 | 54.53 | 55.44 | 69.62 | 41.78 | 63.76 |
| SVM-comgfk | 74.47 | 70.15 | 59.78 | 75.09 | 73.99 | 54.59 | 55.88 | 70.23 | 41.85 | 64.00 |
| ML-rbf | 42.25 | 73.69 | 75.53 | 66.75 | 77.51 | 54.43 | 33.50 | 23.57 | 34.92 | 53.57 |
| ML-comgfk | 80.25 | 74.99 | 78.79 | 67.41 | **77.82** | **71.68** | 49.96 | 50.79 | 53.79 | 67.28 |
| ELM-rbf | 70.63 | 66.44 | 66.83 | 63.45 | 69.73 | 51.23 | 49.76 | 49.83 | 33.50 | 57.93 |
| OSC | **88.10** | 66.71 | 54.66 | 53.81 | 65.13 | 63.71 | 36.05 | 40.21 | 40.08 | 56.5 |
| GLSW | 78.38 | 69.36 | **80.75** | 74.62 | 69.43 | 44.28 | 48.64 | 67.87 | 46.58 | 64.43 |
| DS | 69.37 | 46.28 | 41.61 | 58.88 | 48.83 | 32.83 | 23.47 | **72.55** | 29.03 | 46.98 |
| DRCA | 66.24 | 71.82 | 48.45 | 85.28 | 69.87 | 50.18 | 53.74 | 69.15 | 44.61 | 62.15 |
| D-DRCA | 84.32 | **90.10** | 67.08 | **91.37** | **84.48** | 60.89 | 65.65 | 70.85 | **49.50** | **73.80** |

**Table 5.** Complex E-nose dataset from CQU with dimension of features (DoF) and sample count of different types of gases.

| | DoF | Ammonia | Benzene | Carbon monoxide | Formaldehyde | Nitrogen dioxide | Toluene | Total |
|---|---|---|---|---|---|---|---|---|
| Master | 6 | 60 | 72 | 58 | 126 | 38 | 66 | 420 |
| Slave 1 | 6 | 81 | 108 | 98 | 108 | 107 | 106 | 608 |
| Slave 2 | 6 | 84 | 87 | 95 | 108 | 108 | 94 | 576 |

**Table 6.** Recognition accuracy of the sensor drift dataset from CQU. Bold values represent the best results. The proposed D-DRCA method outperforms the competing ones in both tasks.

| Task | SVM | PCA | LDA | GLSW | DS | DRCA | D-DRCA |
|---|---|---|---|---|---|---|---|
| master→slave 1 | 45.89 | 46.22 | 42.11 | 41.45 | 40.30 | 61.18 | **64.14** |
| master→slave 2 | 31.08 | 41.84 | 41.32 | 48.09 | 39.76 | 59.55 | **66.32** |
| Average | 38.49 | 44.03 | 41.72 | 44.77 | 40.03 | 60.37 | **65.23** |